\documentclass[11pt,a4paper]{article}
\usepackage[hyperref]{acl2017}
\usepackage{times}
\usepackage{latexsym}
\usepackage[utf8x]{inputenc}

\usepackage{url}
\aclfinalcopy

\usepackage{multirow}
\usepackage{placeins}
\usepackage{arydshln}
\usepackage{subcaption}
\usepackage{color}
\usepackage[]{todonotes}

\usepackage[T1]{fontenc}
\usepackage[acronym,shortcuts,smallcaps]{glossaries}
\glsdisablehyper

\newacronym{NMT}{NMT}{neural machine translation}
\newacronym{MT}{MT}{machine translation}
\newacronym{APE}{APE}{automatic post-editing}
\newacronym{BT}{BT}{back-translated}
\newacronym{PBSMT}{PBSMT}{phrase-based statistical machine translation}
\newacronym{SGD}{SGD}{stochastic gradient descent}
\newacronym{SMT}{SMT}{statistical machine translation}
\newacronym{BPE}{BPE}{Byte-Pair Encoding}

\usepackage{adjustbox}

\title{Combining SMT and NMT Back-Translated Data for Efficient NMT}

\author{
\begin{tabular}{c}
\shortstack{Alberto Poncelas, Maja Popović, Dimitar Shterionov,}\\
\shortstack{Gideon Maillette de Buy Wenniger \and Andy Way}\\
\end{tabular}
\\
School of Computing, DCU, ADAPT Centre\\
{\tt \{firstname.lastname\}@adaptcentre.ie}
}

\date{}

\begin{document}
\maketitle
\begin{abstract}
Neural Machine Translation (NMT) models achieve their best performance when large sets of parallel data are used for training. Consequently, techniques for augmenting the training set have become popular recently. One of these methods is back-translation \citep{sennrich2015improving}, which consists on generating synthetic sentences by translating a set of monolingual, target-language sentences using a Machine Translation (MT) model.

Generally, NMT models are used for back-translation. In this work, we analyze the performance of models when the training data is extended with synthetic data using different MT approaches. In particular we investigate back-translated data generated not only by NMT but also by Statistical Machine Translation (SMT) models and combinations of both. The results reveal that the models achieve the best performances when the training set is augmented with back-translated data created by merging different MT approaches.

\end{abstract}


\section{Introduction}
Machine translation (MT) nowadays is heavily dependent on the quantity and quality of training data. The amount of available good-quality parallel data for the desired domain and/or language pair is often insufficient to reach the required translation performance. In such  cases,  it  has  become  the  norm  to resort to back-translating freely available monolingual data as proposed in \cite{sennrich2015improving}. That is, one can translate a set of sentences from language L2 into L1 with an already trained MT system for the language pair L2$\rightarrow$L1. Then create a synthetic parallel corpus from L1 to L2, with the source (L1) side being the translated text and the target side being the monolingual data. Back-translation has been shown to be beneficial not only for MT but also for other NLP tasks where data is scarce, e.g. \ac{APE} \cite{Junczys-Dowmunt2016a,Negri2018}. However, the effects of various parameters for creating \ac{BT} data have not been investigated enough as to indicate what are the optimal conditions in not only creating but also employing such data to train high-quality \ac{NMT} systems. 

The work presented in \citet{bt-eamt18} draws an early-stage empirical roadmap to investigating the effects of \ac{BT} data. In particular, it looks at how the amount of \ac{BT} data impacts the performance of the final NMT system. In \citet{sennrich2015improving} and \citet{bt-eamt18}, the systems used to generate the \ac{BT} data are neural. However, it has been noted that often different paradigms can contribute differently to a given task. For example, it has been shown that applying an \ac{APE} system based on NMT technology improves \ac{SMT} output, but has lower impact on \ac{NMT} output \cite{bojar2017findings,Chatterjee2018}. 

In this work we assess the impact of different amounts of \ac{BT} data generated by two different types of MT systems -- \ac{NMT} and \ac{SMT}. Our contribution is two-fold: (i) we provide a systematic comparison of the \ac{BT} data by building \ac{NMT} systems with a combination of \ac{SMT} and \ac{NMT} \ac{BT} data and (ii) we identify the effects of \ac{BT} data that originates from SMT or NMT on the end-quality of the trained NMT system. We aim to answer the question: "What is the best choice for \ac{BT} data?"


\section{Preparatory Study: the Effect of Back-Translation when Controlling for the Amount of Training Effort}
A typical assumption made when training \ac{NMT} models, is that when more training data is used, more training effort is warranted. Based on this assumption when training NMT systems what is normally kept constant is the amount of training epochs rather than the amount of training effort in the form of steps/mini-batches. Nevertheless, when adding back-translated data to the training set, while keeping the amount of epochs the same, the effective amount of training increases. 
It could then be questioned whether the extra training effort in itself does not partly explain the positive effect of back-translation. For this reason, we seek to answer the question: ``Does the effect of back-translation change when we control for the amount of training effort, by keeping the total amount of steps/mini-batches constant?".
To answer this question we compare the performance of systems trained on purely authentic data to those trained on authentic plus synthetic data, while keeping either the number of steps/mini-batches or the number of epochs constant in both settings:
\begin{enumerate}
    \item Models trained with 1M auth + 2M synth sentences using the default settings, including 13 training epochs.
    \item Models trained on 1M auth data only, trained either: 
    \begin{enumerate}
        \item using the default settings, including 13 training epochs.
        \item Trained for 39 epochs, to obtain a same amount of training effort as for the 1M auth + 2M synth sentences model.
    \end{enumerate}    
\end{enumerate}  
When increasing the epochs to 39, we take appropriate measures to keep the starting point and speed of decay of the learning rate constant for the amount of training steps/epochs.\footnote{This is implemented by changing the start of the learning rate decay from epoch 8 to epoch 22 $(= 7* 3 + 1)$ and changing the decay factor from 0.5 to $\sqrt[3]{0.5} = 0.7936$. This way, the learning rate decay starts after the same amount of data when using the 1M auth dataset $(7 \times 3M)$ and the decay rate is maintained at 0.5 for each $3M$ sentences from this point onwards.}

The results of these experiments indicate that training a model on authentic data with $1/3$ of the amount of the total parallel data (authentic + synthetic) for an additional 26 epochs to account for the extra training effort is not required as no significant improvement has been observed. Based on the outcome of these experiments we chose the rest of our experiments.

\section{Using Back-Translation from Different Sources}\label{sec:different_sources}

The work of \citep{sennrich2015improving} showed that adding \ac{BT} data is beneficial to achieve better translation performances. In this work we compare the details related to the translation hypotheses originating from \ac{SMT} and \ac{NMT} back-translated training data as well as combine the data from those two different sources. To the best of our knowledge, this has not been investigated yet.

We compare German-to-English translation hypotheses generated by systems trained (i) only on authentic data, (ii) only on synthetic data, and (iii) on authentic data enhanced with different types of \ac{BT} data:  \ac{SMT}, \ac{NMT}. We exploit two types of synthetic and authentic data combinations: (a) randomly selected half of target sentences back-translated by SMT and another half by NMT system, and (b) joining all \ac{BT} data (thus repeating each target segment).

The translation hypotheses are compared in terms of four 
automatic evaluation metrics:  BLEU~\cite{papineni2002bleu}, TER~\cite{snover2006study}, METEOR~\cite{banerjee2005meteor} and CHRF~\cite{popovic2015chrf}.  These metrics give an overall estimate of the quality of the translations with respect 
to the reference (human translation of the test set).
In addition, the translation hypotheses are analyzed in terms of five error categories, lexical variety and syntactic variety.

\section{Related Work}\label{sec:related_work}

A comparison between MT models trained with synthetic and with authentic data that originate from the same source has been presented in~\citet{bt-eamt18}. They show that while the performances of models trained with both synthetic and authentic data are better than those of models trained with only authentic data, there is a saturation point beyond which the quality does not improve by adding more synthetic data. Nonetheless, models trained only with synthetic (BT) data perform very reasonably, with evaluation scores being close to those of models trained with only authentic parallel data. In fact, when appropriately selected, BT data can be used to enhance NMT models \citep{poncelas2019adaptation}.


\citet{D18-1045} confirmed that synthetic data can sometimes match the performance of authentic data. In addition, a comprehensive analysis of different methods to generate synthetic source sentences was carried out. This analysis revealed that sampling from the model distribution or noising beam outputs out-performs pure beam search, which is typically used in NMT. Their analysis shows that synthetic data based on sampling and noised beam search provides a stronger training signal than synthetic data based on argmax inference. 

One of the experiments reported in \citet{W18-6315} is comparing performance between models trained with \ac{NMT} and \ac{SMT} \ac{BT} data. The best Moses system \citep{koehn2007moses} is almost as good as the NMT system trained with the same (authentic) data, and much faster to train.  Improvements obtained with the Moses system trained with a small training corpus are much smaller; this system even decreases the performance for the out-of-domain test.
The authors also investigated some properties of \ac{BT} data and found out that the back-translated sources are on average shorter than authentic ones, syntactically simpler than authentic ones, and contain smaller number of rare events. Furthermore, automatic word alignments tend to be more monotonic between artificial sources and authentic targets than between authentic sources and authentic targets.

\citet{W18-6315} also compared training \ac{BT} data with authentic data in terms of lexical and syntactic variety, segment length and alignment monotony, however they did not analyze the obtained translation hypotheses. In \cite{Vanmassenhove2019} it is shown that MT systems trained on authentic and on backtranslated data lead to general loss of linguistic richness in their translation hypotheses.




\section{Experimental Settings}
\label{sec:experimental_settings}

For the experiments we have built German-to-English NMT models using the Pytorch port of OpenNMT \citep{opennmt}. We use the default parameters: 2-layer LSTM with 500 hidden units. The models are trained for the same number of epochs. As the model trained with all authentic data converges after 13 epochs, we use that many iterations to train the models (we use the same amount of epochs). As optimizer we use \ac{SGD}, in combination with learning rate decay, halving the learning rate starting from the 8th epoch.

In order to build the models, all data sets are tokenized and truecased and segmented with \ac{BPE} \citep{sennrich2016neural} built on the joint vocabulary using 89500 merge operations. For testing the models we use the test set provided in the WMT 2015 News Translation Task  \citep{bojar-EtAl:2015:WMT}. As development set, we use 5K randomly sampled sentences from development sets provided in previous years of WMT.



\section{Data} 
\label{sec:datasets}

The parallel data used for the experiments has been obtained from WMT 2015 \citep{bojar-EtAl:2015:WMT}. We build two parallel sets with these sentences: \textit{base} (1M sentences) and \textit{auth} (3M sentences). We use the target side of \textit{auth} to create the following datasets:


\begin{itemize}
    \item \textit{SMTsynth}: Created by translating the target-side sentences of \textit{auth}. The model used to generate the sentences is an \ac{SMT} model  trained with \textit{base} set in the English to German direction. It has been built using the Moses toolkit with default settings, using GIZA++ for word alignment and tuned using MERT \citep{och2003minimum}). The language model (of order 8) is built with the KenLM toolkit \cite{heafield2011kenlm} using the German side of \textit{base}.
    \item \textit{NMTsynth}: Created by translating the target-side sentences of \textit{auth}. The model used to generate the sentences is an \ac{NMT} model (with the same configuration as described in Section \ref{sec:experimental_settings} but in the English to German direction) trained with the \textit{base} set.
    \item \textit{hybrNMTSMT}: Synthetic parallel corpus combining \textit{NMTsynth} and \textit{SMTsynth} sets. It has been built by maintaining the same target side of \textit{auth}, and as source side we alternate between \textit{NMTsynth} and \textit{SMTsynth} each 500K sentences.
    \item \textit{fullhybrNMTSMT}: Synthetic parallel corpus combining all segments from \textit{NMTsynth} and \textit{SMTsynth} sets (double size, each original target sentence repeated twice with both an NMT and SMT back-translation-generated translation).
\end{itemize}


\section{Experiments}


In our experiments, we build models on different portions of the datasets described in Section \ref{sec:datasets}. First, we train an initial NMT model using the \textit{base} data set. Then, in order to investigate how much the models benefit from using synthetic data generated by different approaches, we build models with increasing sizes of data (from the data sets described in Section \ref{sec:datasets}).

The models explored are built with data that ranges from 1M sentences (built with only authentic data from \textit{base} data set) to 4M sentences (consisting on 1M sentences from \textit{base} and 3M sentences generated artificially with different models). We also include the models built with the \textit{fullhybrNMTSMT} set. As this set contains duplicated target-side sentences, the largest model we build contains 7M sentences in total but only 4M distinct target-side sentences.


\section{Results}

\subsection{Controlling the Amount of Training Effort}
Table \ref{table:results-experiments-with-controlled-amount-of-training-effort} shows the effect of controlling the amount of training effort when using back-translation.
It can be observed that increasing the number of epochs from 13 to 39 when using just the 1M base training set does not increase the performance over using just 13 epochs (i.e. not compensating the relatively smaller training set with more epochs), rather it deteriorates it. 
From these results we conclude that there is no reason to believe that the positive effects of using back-translation is caused by an effectively larger training effort, rather than by the advantage of the larger training set itself. 
We therefore also conclude that it is reasonable to keep the number of epochs constant across experiments, rather than fixing the amount of training effort as measured by steps/mini-batches, and we do the former throughout the rest of the paper.


\begin{table*}[ht]
\centering
\begin{tabular}{|p{2cm}|p{2.5cm}|p{2.5cm}|p{2.5cm}|}
\hline
&	 1M \textit{base}.- 13 Epochs  	&	1M \textit{base}.- 39 Epochs-  	&	1M \textit{base} + 2M \textit{NMTsynth}	\\
\hline
 BLEU$\uparrow$	&	23.40	&	23.22	&	25.44		\\
 TER$\downarrow$ 	&	57.23	&	58.21	&	55.62		\\
 METEOR$\uparrow$	&	28.09	&	27.75	&	29.47		\\
CHRF1$\uparrow$	& 50.66&	50.18	&	52.5		\\
\hline											

\end{tabular}
\caption{Results for experimental procedure validation: checking that it is reasonable to use constant number of epochs, not constant amount of training effort, in the experiments.} \label{table:results-experiments-with-controlled-amount-of-training-effort}
\end{table*}


\begin{table*}[!htbp]
\centering
\begin{tabular}{|p{0.25cm}|p{1.6cm}|p{2cm}|p{2.2cm}|p{2.2cm}|p{2.2cm}||p{2.2cm}|}
\hline
&	&	 1M \textit{base}.  	&	-  	&	-  	&	-	&	-	\\
\hline
\multirow{4}{*}{\rotatebox[origin=c]{90}{\centering 1M lines}}	
& BLEU$\uparrow$	&	23.40	&	-	&	-	&	-	&	-	\\
& TER$\downarrow$ 	&	57.23	&	-	&	-	&	-	&	-	\\
& METEOR$\uparrow$	&	28.09	&	-	&	-	&	-	&	-	\\
&CHRF1$\uparrow$	&	50.66	&	-	&	-	&	-	&	-	\\
\hline											
&	&	  + 1M \textit{auth}  	&	 + 1M \textit{SMTsynth}	&	  +1M \textit{NMTsynth} 	&	 + 1M \textit{hybrNMTSMT}	&	 + 2M \textit{fullhybrNMTSMT}	\\
\hline											
\multirow{4}{*}{\rotatebox[origin=c]{90}{\centering 2M lines}}											
& BLEU$\uparrow$	&	24.87	&	24.38	&	25.32	&	25.21	&	25.34	\\
& TER$\downarrow$ 	&	55.81	&	56.05	&	55.66	&	55.87	&	55.79	\\
& METEOR$\uparrow$	&	29.16	&	28.93	&	29.33	&	29.29	&	29.47	\\
&CHRF1$\uparrow$	&	52.03	&	51.89	&	52.25	&	52.36	&	52.47	\\
\hline											
&	&	  + 2M \textit{auth}.  	&	  + 2M \textit{SMTsynth}	&	 + 2M \textit{NMTsynth}	&	 + 2M \textit{hybrNMTSMT}	&	  + 4M \textit{fullhybrNMTSMT}	\\
\hline											
\multirow{4}{*}{\rotatebox[origin=c]{90}{\centering 3M lines}}											
& BLEU$\uparrow$	&	25.69	&	24.58	&	25.44	&	25.62	&	25.94	\\
& TER$\downarrow$ 	&	54.99	&	55.7	&	55.62	&	55.25	&	55.11	\\
& METEOR$\uparrow$	&	29.7	&	29.02	&	29.47	&	29.73	&	29.97	\\
& CHRF1$\uparrow$	&	52.77	&	52.09	&	52.5	&	52.89	&	53.11	\\
\hline											
&	&	+ 3M \textit{auth}	&	 + 3M \textit{SMTsynth}	&	  + 3M \textit{NMTsynth}	&	 +3M  \textit{hybrNMTSMT}	& + 6M \textit{fullhybrNMTSMT}	\\
\hline											
\multirow{4}{*}{\rotatebox[origin=c]{90}{\centering 4M lines}}											
& BLEU$\uparrow$	&	25.97	&	24.65	&	26.01	&	25.83	&	25.86	\\
& TER$\downarrow$ 	&	54.54	&	55.58	&	55.33	&	55.17	&	54.95	\\
& METEOR$\uparrow$	&	29.91	&	29.26	&	29.71	&	29.74	&	29.88	\\
&CHRF1$\uparrow$	&	53.16	&	52.24	&	52.87	&	52.84	&	53.11	\\
\hline							
\end{tabular}
\caption{Performance of models built with increasing sizes of authentic set (first column) and different synthetic datasets (last four columns). +1M, +2M and +3M indicate the amount of sentences added to the \textit{base} set (1M authentic sentences).
\label{table:results}}
\end{table*}



\subsection{Addition of Synthetic Data from SMT and NMT Models }

Table \ref{table:results} shows the results of the performance of the different \ac{NMT} models we have built. The sub-tables indicate the size of the data used for building the models (from 1M to 4M lines). In each column it is indicated whether \textit{base} has been augmented with the \textit{auth}, \textit{SMTsynth}, \textit{NMTsynth}, \textit{hybrNMTSMT}, or \textit{fullhybrNMTSMT} data set. 


The results show that adding synthetic data has a positive impact on the performance of the models as all of them achieve improvements when compared to that built only with authentic data \textit{1M base}. These improvements are statistically significant at p=0.01 (computed with multeval \citep{clark2011better} using Bootstrap Resampling \citep{koehn04}). However, the increases of quality are different depending on the approach followed to create the \ac{BT} data.



First, we observe that models in which SMT-generated data is added do not outperform the models built with the same size of authentic data. For example, the models built with 4M sentences (1M authentic and 3M SMT-produced sentences, in cell \textit{+ 3M SMTsynth}) achieve a performance comparable to the model trained with smaller number of sentences of authentic data (such as \textit{ + 1M auth} cell, 2M sentences).


Models built by using \ac{NMT}-created data have a better performance than those built with data generated by \ac{SMT}. When performing a pair-wise comparison between models using an equal amount of either \ac{SMT} or \ac{NMT}-created data, we observe that the latter models outperform the former by around one BLEU point.
In fact, the performance of models using NMT-translated sentences is closer to those built with authentic data, and some \textit{NMTsynth} models produce better translation qualities. This is the case of \textit{+1M NMTsynth} model (according to all evaluation metrics) or \textit{+3M NMTsynth} (according to BLEU). 





Our experiments also include the performance of models augmented with a combination of SMT- and NMT- generated data. We see that adding \textit{hybrNMTSMT} data, with one half of the data originating from SMT and the other half from NMT models, have performances similar to those models built on authentic data only. According to some evaluation metrics, such as METEOR, the performance is better than \textit{auth} models when adding 1M or 2M artificial sentences (although none of these improvements are statistically significant at p=0.01). For these amount of sentences, it also outperforms those models in which only SMT or only NMT \ac{BT} data have been included.

The models extended with synthetic data that perform best are \textit{fullhybrNMTSMT} models. Furthermore, they also outperform authentic models when built with less than 4M distinct target-sentences according to BLEU, METEOR (showing statistically significant improvements at p=0.01) and CHRF1. Despite that, when using large sizes of data (i.e. adding 3M synthetic sentences) the models built with SMT-generated artificial data have the lowest performances whereas the performance of the other three tends to be similar.

\subsection{Further Analysis}

In order to better understand the described systems, we carried out more detailed analysis of all translation outputs. We analyzed five error categories: morphological errors, word order, omission, addition and lexical errors, and we compared lexical and syntactic variety of different outputs in terms of vocabulary size and number of distinct POS n-grams. We also analyzed the sentence lengths in different translation hypotheses, however no differences were observed, neither in the average sentence length nor in the distribution of different lengths. 
 
\subsubsection*{Automatic Error Analysis}

\begin{table*}[ht]
    \centering
    \begin{tabular}{|l|ccccc|} \hline
                    & \multicolumn{5}{c|}{error class rates$\downarrow$} \\
       training  &  morph & order & omission & addition & mistranslation \\ \hline
       1M \textit{base} & 2.8  & 9.8  & 12.0 & 4.8 & 29.1 \\ \hline
       1M \textit{base} + 1M \textit{auth} & 2.7  & 9.5  & 11.4  & 4.9  & 28.2 \\ \hdashline
       1M \textit{base} + 1M \textit{SMTsynth} & 2.8  & 10.0  & 11.6  &  4.8  & 28.1  \\
       1M \textit{base} + 1M \textit{NMTsynth} & 2.7  & 9.8  & 10.9  & 5.0  & 28.1 \\
       1M \textit{base} + 1M \textit{hybrNMTSMT} & 2.7 & 9.6  & 11.4  & 5.2  & 27.7 \\
       1M \textit{base} + 1M \textit{fullhybrNMTSMT}  & 2.6  & 9.5  & 11.0 & 5.2 & 27.8  \\ \hline
       1M \textit{base} + 2M \textit{auth} & 2.6 & 9.6 & 11.2 & 4.8 & 27.7\\ \hdashline
       1M \textit{base} + 2M \textit{SMTsynth} & 2.7 & 10.0 & 11.9 & 4.5 & 28.0 \\
       1M \textit{base} + 2M \textit{NMTsynth} & 2.6 & 9.7 & 11.1 & 5.1 & 27.9 \\
       1M \textit{base} + 2M \textit{hybrNMTSMT} & 2.6 & 9.6 & 11.0 & 5.2 & 27.6 \\
       1M \textit{base} + 2M  \textit{fullhybrNMTSMT} & 2.6 & 9.6 & 10.7 & 5.3 & 27.4 \\ \hline
       1M \textit{base} + 3M \textit{auth} & 2.7 & 9.8 & 11.2 & 4.6 & 27.6 \\ \hdashline
       1M \textit{base} + 3M \textit{SMTsynth}& 2.7 & 9.8 & 11.9 & 4.6 & 27.9 \\
       1M \textit{base} + 3M \textit{NMTsynth} & 2.5 & 9.6 & 11.3 & 5.3 & 27.4 \\
       1M \textit{base} + 3M \textit{hybrNMTSMT} & 2.6 & 9.5 & 11.0 & 5.1 & 27.6 \\
       1M \textit{base} + 3M  \textit{fullhybrNMTSMT} & 2.5 & 9.7 & 10.8 & 4.8 & 27.7 \\ \hline
    \end{tabular}
    \caption{Results of automatic error classification into five error categories: morphological error (morph), word order error (order), omission, addition and mistranslation.}
    \label{tab:auterr}
\end{table*}

For automatic error analysis results, we used  Hjerson~\citep{hjerson11}, an open-source tool based on Levenshtein distance, precision and recall. The results are presented in Table~\ref{tab:auterr}.

It can be seen that morphological errors are slightly improved by any additional data, but it is hard to draw any conclusions. This is not surprising given that our target language, English, is not particularly morphologically rich. Nevertheless, for all three corpus sizes, the numbers are smallest for the full hybrid system, being comparable to the results with adding authentic data. 

As for word order, adding SMT data is not particularly beneficial since it either increases (1M and 2M) or does not change (3M) this error type. NMT systems alone do not help much either, except a little bit for the 3M corpus. Hybrid systems yield the best results for this error category for all corpus sizes, reaching or even slightly surpassing the result with authentic data. 

Furthermore, all \ac{BT} data are beneficial for reducing omissions, especially hybrid which can be even better than the authentic data result. 

As for additions, no systematic changes can be observed, except an increase for all types of \ac{BT} data. However, it should be noted that this error category is reported not to be very reliable for comparing different MT outputs~(see for example \cite{humaut11}).

The mostly affected error category is mistranslations. All types of additional data are reducing this type of errors, especially the hybrid \ac{BT} data for 1M and 2M, even surpassing the effect of adding authentic data. As for the 3M corpus, the improvement in this error category is similar to the one by authentic data, but the best option is to use NMT \ac{BT} data alone.

In total, the clear advantage of using hybrid systems can be noted for mistranslations, omissions and word order which is the most interesting category. This error category is augmented by adding BT SMT data or not affected by adding BT NMT data, but combining two types of data creates beneficial signals in the source text.

\subsubsection*{Lexical and Syntactic Variety}

\begin{figure*}[ht]
\includegraphics[width=15cm, height=6cm]{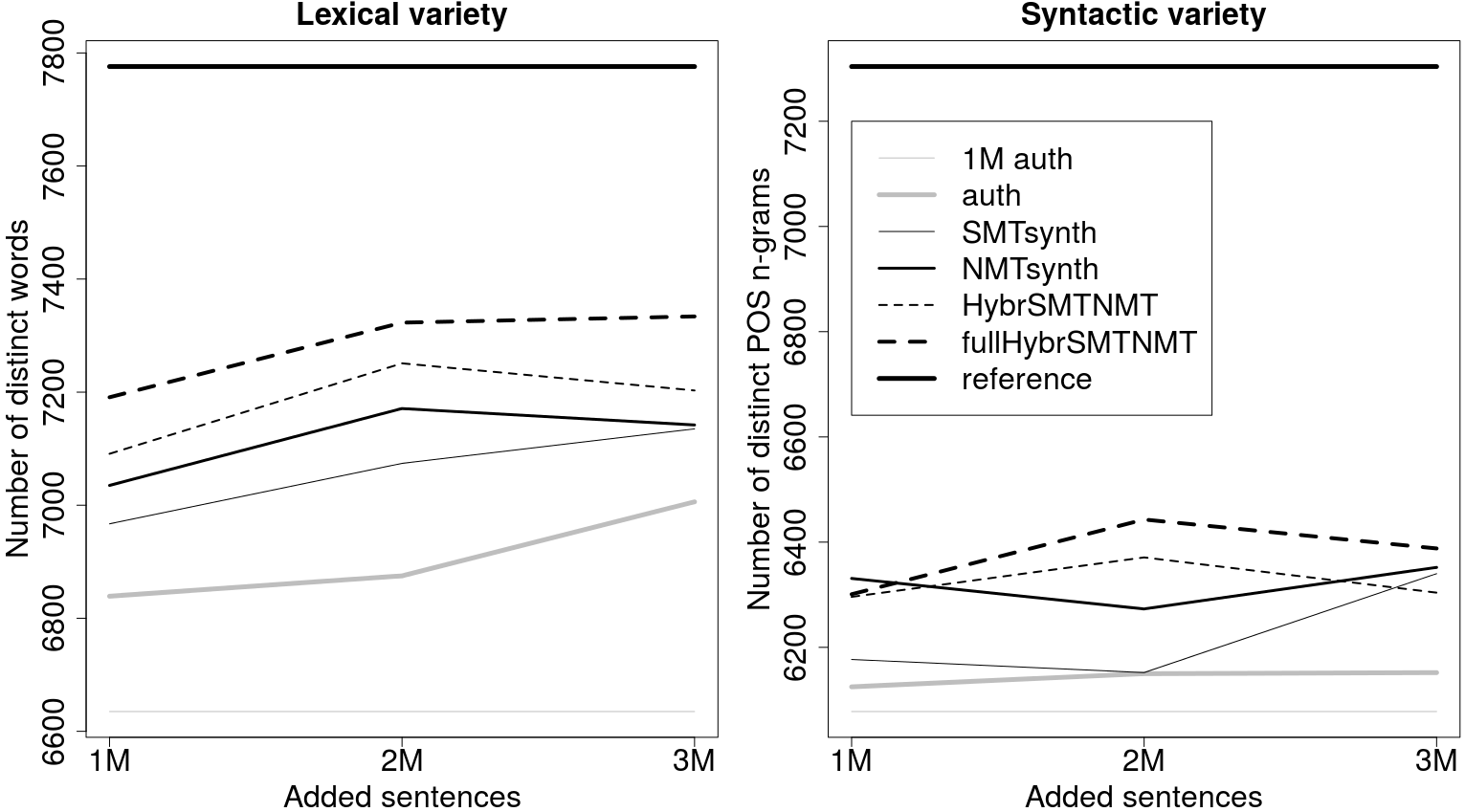}
\centering
\caption{Lexical variety and syntactic variety for all translation hypotheses and for human reference translations.}
\label{fig:variety}
\end{figure*}

\begin{figure*}[ht]
\includegraphics[width=15cm, height=9.5cm]{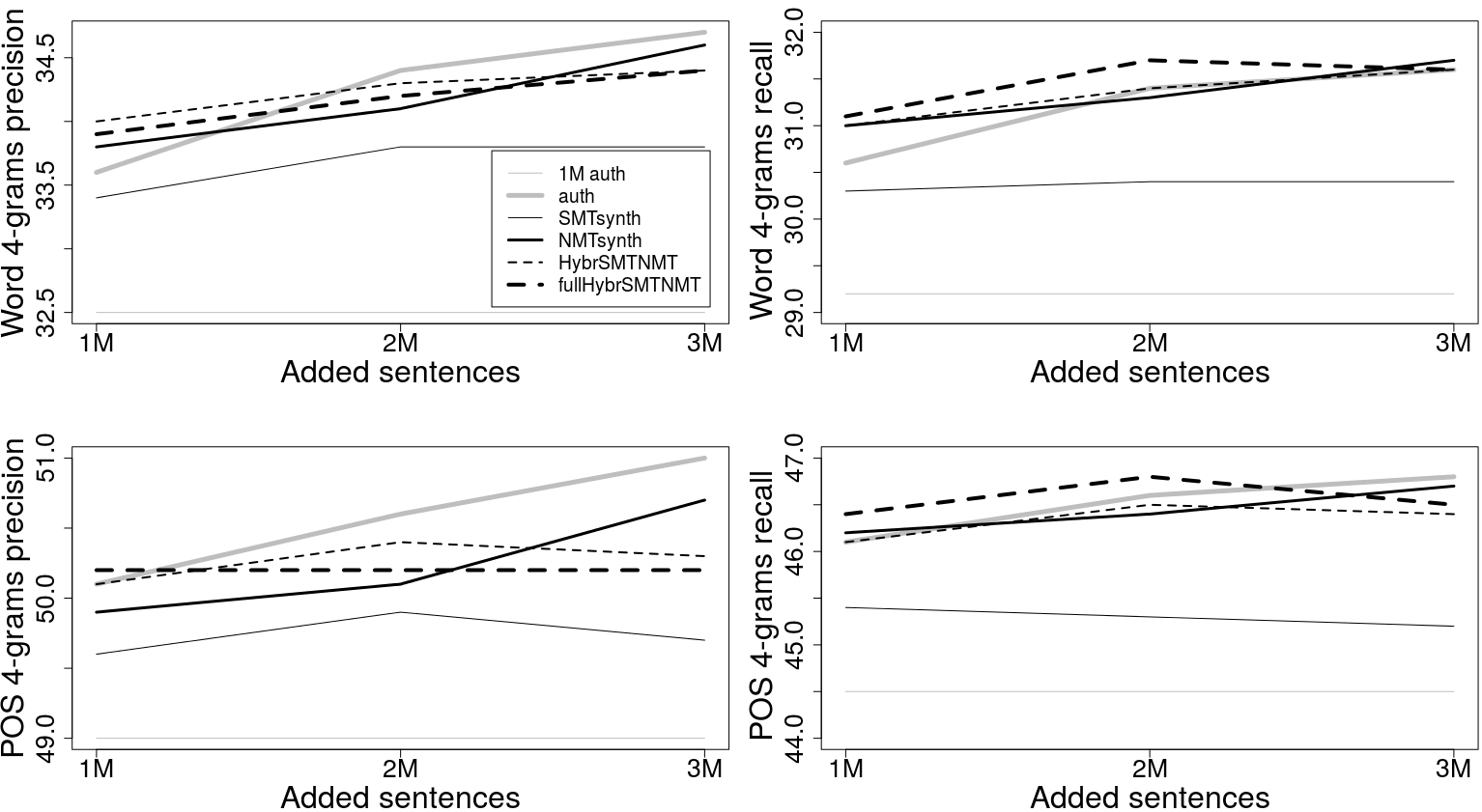}
\centering
\caption{Word/POS 4-gram precision and recall for all translation hypotheses.}
\label{fig:precrec}
\end{figure*}

Lexical and syntactic variety is estimated for each translation hypothesis as well as for the human reference translation. The motivation for this is the observation that machine-translated data is generally lexically poorer and syntactically simpler than human translations or texts written in the original language~\cite{Vanmassenhove2019}. We want to see how different or similar our translation hypotheses are in this sense, and also how they relate to the reference. 

Lexical variety is measured by vocabulary size (number of distinct words) in the given text, and syntactic variety by number of distinct POS n-grams where $n$ ranges from 1 to 4. The results are shown in Figure~\ref{fig:variety}.

First of all, it can be seen that none of the translation hypotheses reaches the variety of the reference translation (the black line on the top). The difference is even more notable for the syntax, where the differences between translation hypotheses are smaller  and the difference between them and the reference is larger than for vocabulary. 

Furthermore, it can be seen that for authentic data (thin gray line on the bottom and thick gray line) the variety increases monotonically with adding more text. 

Lexical variety is increased by all synthetic data, too, even more than by authentic data, however, for the NMT and hybrid synthetic data the increase for the 3M corpus is smaller than for smaller corpora. 

The increase of syntactic variety is lower both for authentic and for synthetic data than the increase of lexical variety. For 1M and 2M corpus, syntactic variety is barely increased by SMT synthetic data whereas NMT and hybrid data are adding more new instances. For the 3M corpus, however, all synthetic methods yield similar syntactic variety, larger than the one obtained by adding authentic data. 



\subsubsection*{Word/POS 4-gram Precision and Recall}
Whereas the increase of lexical and syntactic varieties is a positive trend in general, there is no guarantee that the MT systems are not introducing noise thereby. To estimate how many of added words and POS sequences are sensible, we calculate precision and recall of word and POS 4-grams when compared to the given reference translation. The idea is to estimate how much the translation hypotheses are getting closer to the reference. We take word 4-grams instead of single words because it is not only important that a word makes sense in isolation, but also in a context. Of course, it is still possible that some of the new instances are valid despite being different from the given single reference. 



The results of precision and recall for word/POS 4-grams are
are shown in Figure~\ref{fig:precrec}. 
Several tendencies can be observed:
\begin{itemize}
    \item hybrid \ac{BT} data is especially beneficial for the 1M and 2M additional corpora, for 1M even outperforming the authentic additional data, especially regarding word 4-grams; 
    
    \item NMT BT is the best synthetic option for the 3M additional corpus, however not better than adding 3M of authentic data. This tendency is largest for POS 4-gram precision.
    
    \item SMT \ac{BT} data achieves the lowest scores, especially for POS 4-grams; this is probably related to the fact that it produces less grammatical BT sources, which are then propagated to the translation hypotheses. The differences are largest for the 3M additional corpus, which is probably the reason of diminished effect of the hybrid \ac{BT} data for this setup.  
\end{itemize}

Overall tendencies are that the hybrid \ac{BT} data is capable even of outperforming the same amount of authentic data if the amount of added data does not exceed the double size of the baseline authentic data. For larger data, a deterioration can be observed for the SMT \ac{BT} data, leading to saturation of hybrid models. 

Further work dealing with mixing data techniques is necessary, in order to investigate refined selection methods (for example, removing SMT segments which introduce noise). 

\section{Conclusion and Future Work}

In this work we have presented a comparison of the performance of models trained with increasing size of back-translated data. The artificial data sets explored include sentences generated by using an \ac{SMT} model, and \ac{NMT} model and a combination of both. Two mixing strategies are explored: randomly selecting one half of the source segments from the SMT \ac{BT} data and the other half from the NMT \ac{BT} data, and using all BT source segments thus repeating each target segment.

Some findings from previous work~\cite{W18-6315} are confirmed, namely that in terms of overall automatic evaluation scores, SMT \ac{BT} data reaches slightly worse performance than NMT \ac{BT} data. 
Our main findings are that mixing SMT and NMT \ac{BT} data further improves over each data used alone, especially if full hybridisation is used (using two sources for each target side). These data can even reach better performance than adding the same amount of authentic data, mostly by reducing the number of mistranslations, and increasing the lexical and syntactic variety in a positive way (introducing useful new instances). 

However, if the amount of synthetic data becomes too large (three times larger than the authentic baseline data), the benefits of hybrid system start to diminish. The most probable reason is the decrease in grammaticality introduced by SMT \ac{BT} data which becomes dominant for the larger synthetic corpora.

The presented findings offer several directions for the future work, such as exploring efficient strategies for mixing SMT and NMT data for different authentic/synthetic ratios and investigating morphologically richer target languages.

\section*{Acknowledgements}
This research has been supported by the ADAPT Centre for Digital Content Technology which is funded under the SFI Research Centres Programme (Grant 13/RC/2106) and is co-funded under the European Regional Development Fund.

\noindent 
\includegraphics[width=0.9cm]{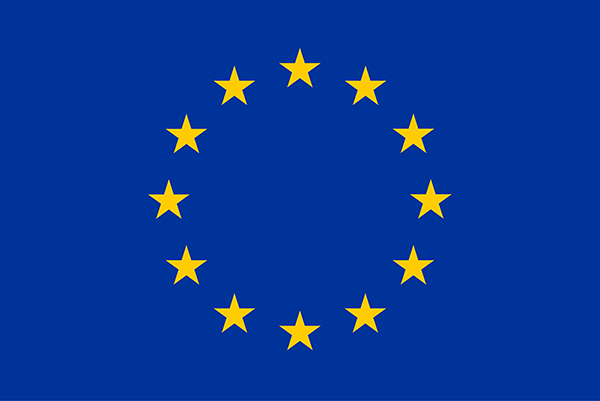}
This work has also received funding from the European Union’s Horizon 2020 research and innovation programme under the Marie Skłodowska-Curie grant agreement No 713567.

\bibliographystyle{acl_natbib}
\bibliography{bibl}

\end{document}